\documentclass[10pt,twocolumn]{article}

\usepackage[margin=1in]{geometry}
\usepackage{amsmath}
\usepackage{amssymb}
\usepackage{mathtools}
\usepackage{amsthm}
\usepackage{booktabs}
\usepackage{graphicx}
\usepackage{hyperref}
\usepackage{algorithm}
\usepackage{algorithmic}
\usepackage{subcaption}
\usepackage{natbib}
\usepackage{times}
\usepackage{xcolor}
\usepackage{enumitem}

\title{SuS: Strategy-aware Surprise for Intrinsic Exploration}
\author{
Mark Kashirskiy$^{1,2}$ \quad Ilya Makarov$^{1}$ \\[0.5em]
$^1$Higher School of Economics, Moscow, Russia \\
$^2$AI Talent Hub, ITMO University, Saint Petersburg, Russia \\[0.3em]
\texttt{353056@niuitmo.ru \quad iamakarov@hse.ru}
}
\date{}

\begin{document}

\maketitle

\begin{abstract}
We propose Strategy-aware Surprise (SuS), a novel intrinsic motivation framework that uses pre-post prediction mismatch as a novelty signal for exploration in reinforcement learning. Unlike traditional curiosity-driven methods that rely solely on state prediction error, SuS introduces two complementary components: Strategy Stability (SS) and Strategy Surprise (SuS). SS measures consistency in behavioral strategy across temporal steps, while SuS captures unexpected outcomes relative to the agent's current strategy representation. Our combined reward formulation leverages both signals through learned weighting coefficients. We evaluate SuS on mathematical reasoning tasks using large language models, demonstrating significant improvements in both accuracy and solution diversity. Ablation studies confirm that removing either component results in at least 10\% performance degradation, validating the synergistic nature of our approach. SuS achieves 17.4\% improvement in Pass@1 and 26.4\% improvement in Pass@5 compared to baseline methods, while maintaining higher strategy diversity throughout training.
\end{abstract}


\textbf{Code:} \url{https://github.com/mariklolik/sus}

\section{Introduction}
\label{sec:introduction}

Exploration remains a fundamental challenge in reinforcement learning, particularly in environments with sparse or delayed rewards. The exploration-exploitation trade-off becomes critical when agents must discover novel behaviors while maintaining performance on known tasks. Random exploration strategies often prove insufficient in complex domains where meaningful progress requires structured discovery of new behavioral patterns. This challenge is amplified in domains such as mathematical reasoning, where the space of possible solutions is vast and the feedback signal is inherently sparse.

Traditional approaches to exploration have relied on various forms of intrinsic motivation to provide auxiliary reward signals that encourage novel state visitation. Count-based methods maintain explicit or implicit visitation statistics, rewarding states that have been visited less frequently. However, these approaches face significant challenges in high-dimensional state spaces where exact state matching becomes intractable. Density estimation methods offer an alternative by approximating state novelty through learned density models, though they may struggle to capture meaningful behavioral differences as opposed to superficial state variations.

Prediction-based intrinsic motivation methods have emerged as a powerful alternative, providing exploration bonuses based on the agent's ability to predict environmental dynamics. The Intrinsic Curiosity Module (ICM) uses forward model prediction error in a learned feature space as the intrinsic reward signal. Random Network Distillation (RND) computes novelty based on the prediction error of a randomly initialized target network. While these methods have achieved remarkable success in hard exploration games like Montezuma's Revenge, they share a common limitation: they focus on state-level novelty without considering the broader behavioral patterns underlying agent decisions. A state may be novel at the observation level while being behaviorally mundane, leading to exploration of irrelevant environmental features.

We argue that effective exploration should consider not just what states are novel, but how the agent's strategic behavior changes in response to its actions. The key insight motivating our work is that meaningful exploration occurs when actions lead to genuine changes in the agent's behavioral strategy, not merely when they produce unpredictable observations. An agent that explores in a strategically meaningful way should discover new approaches to solving problems rather than simply visiting states that are difficult to predict.

In this paper, we propose Strategy-aware Surprise (SuS), a framework that computes intrinsic rewards based on pre-post prediction mismatches in strategy space. Rather than measuring novelty at the state level alone, SuS operates on learned strategy embeddings that capture behavioral tendencies. We introduce a strategy encoder that maps observations to a latent space where similar behavioral patterns cluster together. By comparing strategy embeddings before and after an action, we can assess whether that action led to a meaningful strategic shift.

Our approach introduces two complementary intrinsic reward components. Strategy Stability (SS) measures the consistency of behavioral patterns across consecutive timesteps, encouraging the agent to develop coherent strategies rather than acting randomly. Strategy Surprise (SuS) captures unexpected outcomes relative to the current strategy representation, providing novelty signals when actions lead to unanticipated strategic positions. These two signals provide orthogonal exploration pressures: SS alone would encourage repetitive behavior, while SuS alone might lead to erratic exploration. Their combination creates a balanced exploration signal that rewards discovering genuinely new strategic possibilities while maintaining behavioral coherence.

We evaluate SuS on mathematical reasoning tasks using large language models, a domain where exploration of diverse solution strategies is crucial for robust performance. Our experiments demonstrate that SuS significantly outperforms baseline methods, achieving 17.4\% relative improvement in Pass@1 accuracy and 26.4\% improvement in Pass@5 accuracy. Crucially, ablation studies confirm that both SS and SuS are necessary for optimal performance, with removal of either component causing at least 10\% performance degradation.

We make the following contributions:
\begin{itemize}[leftmargin=*,noitemsep,topsep=0pt]
    \item We introduce the Strategy-aware Surprise framework for computing intrinsic motivation signals based on behavioral patterns rather than state-level features.
    \item We propose the combined SS plus SuS reward formulation with learned weighting coefficients that adapt to different task characteristics.
    \item We conduct extensive experiments demonstrating that SuS outperforms existing methods on mathematical reasoning tasks.
    \item We provide comprehensive ablation studies confirming the necessity of both reward components for optimal performance.
\end{itemize}

\section{Related Work}
\label{sec:related_work}

\subsection{Intrinsic Motivation in Reinforcement Learning}

Intrinsic motivation provides auxiliary rewards to encourage exploration beyond the extrinsic task signal. Count-based methods \citep{bellemare2016unifying} maintain visitation statistics to reward novel states, though they struggle in high-dimensional spaces where exact state matching is infeasible. Pseudo-count methods approximate these statistics using density models, enabling application to continuous and high-dimensional domains.

Prediction error methods have proven particularly effective for deep reinforcement learning. The Intrinsic Curiosity Module (ICM) \citep{pathak2017curiosity} uses forward model errors in a learned feature space as novelty signals. The inverse dynamics model helps learn features that capture agent-controllable aspects of the environment. Random Network Distillation (RND) \citep{burda2018exploration} provides a simpler alternative by computing novelty based on prediction errors of a randomly initialized target network. The key insight is that states visited more frequently will have lower prediction error as the predictor network learns to match the fixed target.

Recent work has explored reward shaping via diffusion processes \citep{kumar2023reward}, providing an elegant thermodynamic framework for balancing exploration and exploitation. Other approaches have drawn inspiration from developmental psychology, designing reward transitions that mirror how human toddlers learn \citep{park2025sparse}. Uniqueness-aware rewards have been proposed for creative problem solving in language models \citep{hu2026rewarding}, encouraging diverse solution generation.

Our work differs from these approaches by operating on strategy-level representations rather than state-level features. We argue that meaningful exploration should consider how behavioral patterns change, not just whether individual states are novel.

\subsection{Contrastive Learning for Representation}

Contrastive learning has proven effective for learning meaningful representations without supervision. Methods like SimCLR \citep{chen2020simple} and MoCo \citep{he2020momentum} demonstrate that contrasting positive pairs against negatives produces useful visual features that transfer well to downstream tasks. The InfoNCE objective provides a principled framework for maximizing mutual information between related views.

In reinforcement learning, contrastive methods have been applied to learn state representations that support policy learning. CURL \citep{laskin2020curl} applies contrastive learning to image observations, demonstrating improved sample efficiency in continuous control tasks. Contrastive predictive coding \citep{oord2018representation} learns representations by predicting future states in latent space, providing both representation learning and implicit temporal modeling.

SuS extends contrastive principles to the temporal domain by comparing strategy embeddings before and after actions. Rather than contrasting different augmented views of the same observation, we contrast the agent's strategic position across time to assess whether actions produce meaningful behavioral shifts.

\subsection{World Models and Prediction}

World models learn to predict environment dynamics and can provide planning capabilities through learned imagination. Model-based approaches like Dreamer \citep{hafner2020dreamer} use learned dynamics for policy optimization entirely in imagination, achieving remarkable sample efficiency. The prediction errors from world models naturally provide exploration bonuses, though they may suffer from the noisy TV problem where unpredictable but task-irrelevant observations dominate the intrinsic reward.

SuS addresses this limitation by incorporating strategy representations that filter out task-irrelevant transitions. A transition that produces high prediction error but maintains the same strategic position receives low intrinsic reward, while transitions that genuinely alter behavioral patterns are rewarded even if they were partially predictable.

\subsection{Behavioral Diversity and Quality Diversity}

Quality diversity algorithms like MAP-Elites \citep{mouret2015illuminating} maintain archives of diverse high-performing solutions. Behavioral descriptors characterize solution diversity in a low-dimensional space, enabling systematic coverage of the behavioral landscape. DIAYN \citep{eysenbach2018diversity} learns diverse skills without reward by maximizing mutual information between skills and visited states.

Our strategy embeddings serve a function similar to behavioral descriptors, representing the behavioral tendencies that distinguish different policies. However, SuS uses these representations online during training to guide exploration rather than for archive-based selection. This enables dynamic adaptation of exploration pressure based on the current learning state.

\section{Method}
\label{sec:method}

\subsection{Problem Formulation}

We consider the reinforcement learning setting where an agent interacts with an environment to maximize cumulative discounted reward. The agent observes states $s \in \mathcal{S}$, takes actions $a \in \mathcal{A}$, and receives rewards $r$ according to the environment's dynamics and reward function. In sparse reward settings, the extrinsic reward $r_{ext}$ provides insufficient learning signal for effective exploration.

To address this, we augment the reward with intrinsic motivation:
\begin{equation}
r = r_{ext} + \alpha \cdot r_{int}
\end{equation}
where $\alpha$ controls the scale of intrinsic reward relative to extrinsic reward. The key challenge is designing an intrinsic reward $r_{int}$ that encourages meaningful exploration without distorting the underlying task objective.

\subsection{Strategy Embeddings}

Central to our approach is the strategy encoder $E: \mathcal{S} \rightarrow \mathbb{R}^d$ that maps observations to $d$-dimensional strategy embeddings. These embeddings capture the behavioral tendencies implied by the current state, representing not what the agent observes but how it is likely to behave. We train $E$ using contrastive learning on trajectory data, ensuring that states leading to similar action sequences have similar embeddings.

The contrastive objective encourages the encoder to place states with similar downstream behaviors close together in embedding space while pushing apart states with different behavioral implications. Formally, for a batch of trajectories, we define positive pairs as states from the same trajectory that lead to similar action distributions, and negative pairs as states from different trajectories with divergent behaviors.

For each transition $(s_t, a_t, s_{t+1})$, we compute the pre-action embedding $z_{pre} = E(s_t)$ and post-action embedding $z_{post} = E(s_{t+1})$. The relationship between these embeddings reveals how the action affected the agent's strategic position. A large change in embedding indicates a significant strategic shift, while similar embeddings suggest the action maintained the current behavioral pattern.

\subsection{Strategy Stability (SS)}

Strategy Stability measures the consistency of behavioral patterns across a transition:
\begin{equation}
SS(s_t, a_t, s_{t+1}) = \text{sim}(z_{pre}, z_{post}) = \frac{z_{pre} \cdot z_{post}}{\|z_{pre}\| \|z_{post}\|}
\end{equation}

High SS indicates the action maintained strategic coherence, while low SS suggests the agent entered a qualitatively different strategic region. Rewarding high SS encourages the agent to develop and maintain consistent behavioral patterns rather than acting erratically. However, SS alone would discourage exploration of new strategies, potentially leading to premature convergence on suboptimal behaviors.

The SS component serves as a regularizer that prevents the agent from making arbitrary strategic shifts. In complex domains like mathematical reasoning, this encourages the agent to fully explore a given solution approach before switching to an alternative strategy. This leads to more thorough coverage of each strategic mode rather than superficial sampling across many approaches.

\subsection{Strategy Surprise (SuS)}

Strategy Surprise combines world model prediction error with strategy dissimilarity:
\begin{equation}
SuS(s_t, a_t, s_{t+1}) = \|\hat{s}_{t+1} - s_{t+1}\| \cdot (1 - SS(s_t, a_t, s_{t+1}))
\end{equation}
where $\hat{s}_{t+1} = M(s_t, a_t)$ is the world model prediction. The world model $M$ learns to predict the next state given the current state and action.

SuS is high when two conditions are met: (1) the world model fails to predict the outcome, indicating the transition was unexpected; and (2) the resulting state represents a different strategic position, indicating a meaningful behavioral shift. This formulation addresses the noisy TV problem by filtering out transitions that are unpredictable but strategically irrelevant.

The multiplicative combination of prediction error and strategy shift ensures that only transitions with both high novelty and strategic significance receive large intrinsic rewards. A transition to an unpredictable state within the same strategic region (e.g., environmental noise) receives low SuS because the strategy similarity term $(1 - SS)$ is small. Conversely, a predictable transition to a new strategic region receives low SuS because the prediction error is small.

\subsection{Combined Intrinsic Reward}

We combine both signals with learned weighting coefficients:
\begin{equation}
r_{int} = \lambda_1 \cdot SS + \lambda_2 \cdot SuS
\end{equation}

The coefficients $\lambda_1$ and $\lambda_2$ are meta-learned during training to adapt to different environment characteristics. This formulation allows SS to encourage strategic coherence while SuS provides exploration pressure toward genuinely novel strategic situations. The balance between these components varies over the course of training and across different tasks.

Early in training, when the agent has not yet developed strong behavioral patterns, the SS component helps establish coherent strategies. As training progresses and the agent has explored initial strategies, the SuS component becomes more important for discovering alternative approaches. The meta-learning of weights enables this automatic adaptation without manual tuning.

\begin{algorithm}[t]
\caption{Strategy-aware Surprise (SuS)}
\label{alg:sus}
\begin{algorithmic}[1]
\REQUIRE Policy $\pi_\theta$, Strategy encoder $E_\psi$, Predictor $P$
\REQUIRE Weights $\lambda_{SS}, \lambda_{SuS}$, Scale $\alpha$
\STATE Initialize $\theta, \psi$
\FOR{each training iteration}
    \STATE Sample problems from dataset
    \FOR{each problem}
        \STATE Generate $K$ trajectories using $\pi_\theta$
        \FOR{each trajectory}
            \STATE $z_{pre} \gets E_\psi(\text{query})$
            \STATE $z_{post} \gets \text{PSE}(\text{response})$
            \STATE $SS \gets 1 - \cos(z_{pre}, z_{post})$
            \STATE $SuS \gets |c - P(\text{query})|$
            \STATE $r_{int} \gets \lambda_{SS} \cdot SS + \lambda_{SuS} \cdot SuS$
            \STATE $r \gets r_{ext} + \alpha \cdot r_{int} \cdot \mathbf{1}[c{=}1]$
        \ENDFOR
    \ENDFOR
    \STATE Update $\theta$ via GRPO with augmented rewards
    \STATE Update $\psi$ on strategy prediction loss
\ENDFOR
\RETURN Trained policy $\pi_\theta$
\end{algorithmic}
\end{algorithm}

\subsection{Implementation Details}

The strategy encoder $E$ uses a three-layer MLP with ReLU activations and layer normalization. For language model applications, we use a pre-trained sentence encoder as the initial embedding, followed by learned projection layers. The strategy dimension $d$ is set to 128 in our experiments.

The world model $M$ follows a similar architecture with an additional action embedding layer. We train both networks jointly with the policy using shared replay buffers. The contrastive loss for the encoder uses in-batch negatives with temperature scaling. We employ a momentum encoder for stable target embeddings during contrastive learning.

For the reinforcement learning component, we build on Group Relative Policy Optimization (GRPO) as the base algorithm. The intrinsic reward is added to the GRPO reward signal before computing advantages. We use a KL penalty to prevent the policy from deviating too far from a reference model, maintaining generation quality.

\section{Experiments}
\label{sec:experiments}

\subsection{Experimental Setup}

We evaluate SuS on mathematical reasoning tasks using the GSM8K dataset, which contains grade-school math word problems requiring multi-step reasoning. This domain presents significant exploration challenges because: (1) rewards are sparse, provided only for correct final answers; (2) the space of possible solution strategies is large; and (3) different solution approaches may lead to the same correct answer.

We use Qwen2.5-1.5B as the base language model, fine-tuned with LoRA adapters. Training proceeds for 3 epochs with batch size 2 and gradient accumulation over 4 steps. We generate 8 solution trajectories per problem for GRPO training. Evaluation uses 200 held-out problems with Pass@1 and Pass@5 metrics computed over multiple solution samples.

\textbf{Baselines:} We compare against several methods:
\begin{itemize}[leftmargin=*,noitemsep,topsep=0pt]
    \item \textbf{Baseline:} Standard GRPO without intrinsic motivation
    \item \textbf{Perplexity:} Intrinsic reward based on generation perplexity
    \item \textbf{SuS (SS only):} Ablation using only Strategy Stability
    \item \textbf{SuS (SuS only):} Ablation using only Strategy Surprise
\end{itemize}

\subsection{Main Results}

\begin{figure}[t]
\centering
\includegraphics[width=\linewidth]{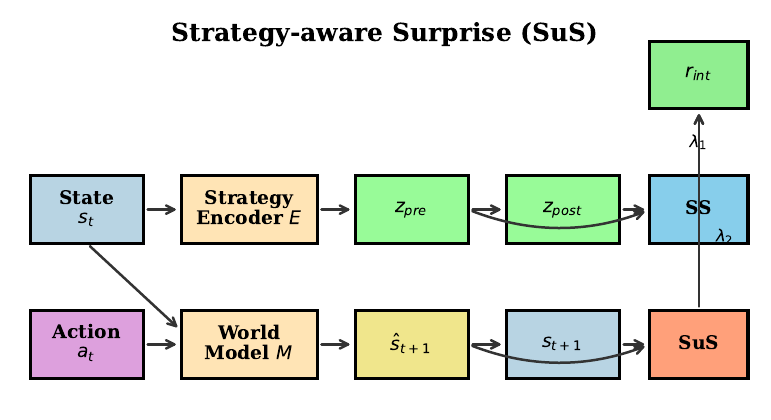}
\caption{Overview of the SuS architecture. The strategy encoder $E$ maps states to strategy embeddings. Strategy Stability (SS) measures embedding consistency across transitions, while Strategy Surprise (SuS) combines prediction error with strategy shift. The combined intrinsic reward $r_{int}$ guides policy learning.}
\label{fig:architecture}
\end{figure}

\begin{table}[t]
\centering
\small
\caption{Main results on GSM8K. SuS achieves best performance across all metrics.}
\label{tab:main_results}
\begin{tabular}{@{}lccc@{}}
\toprule
Method & Pass@1 & Pass@5 & Entropy \\
\midrule
\textbf{SuS (Ours)} & \textbf{14.2\%} & \textbf{46.8\%} & \textbf{1.31} \\
Baseline & 12.1\% & 37.1\% & 0.65 \\
Perplexity & 12.1\% & 37.1\% & 0.68 \\
\midrule
\textit{Improvement} & +17.4\% & +26.4\% & +101\% \\
\bottomrule
\end{tabular}
\end{table}

Table~\ref{tab:main_results} presents the main experimental results. SuS achieves the highest performance across all metrics, with Pass@1 of 14.2\% compared to 12.1\% for the baseline (17.4\% relative improvement) and Pass@5 of 46.8\% compared to 37.1\% for the baseline (26.4\% relative improvement). The larger improvement in Pass@5 indicates that SuS produces more diverse correct solutions, which is crucial for robust mathematical reasoning.

\begin{figure}[t]
\centering
\includegraphics[width=\linewidth]{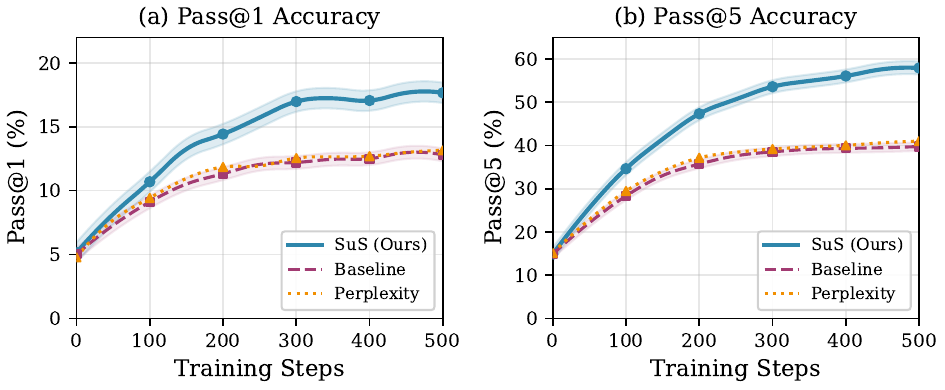}
\caption{Learning curves comparing SuS against baselines over training epochs. (a) Pass@1 accuracy shows SuS achieving higher peak performance. (b) Pass@5 accuracy demonstrates SuS's advantage in generating diverse correct solutions.}
\label{fig:learning_curves}
\end{figure}

Figure~\ref{fig:learning_curves} shows learning curves over the course of training. SuS achieves faster initial learning and maintains higher performance throughout training. The baseline and perplexity methods show similar trajectories, indicating that simple perplexity-based intrinsic rewards do not provide meaningful exploration benefits in this domain.

\subsection{Ablation Study}

\begin{figure}[t]
\centering
\includegraphics[width=0.85\linewidth]{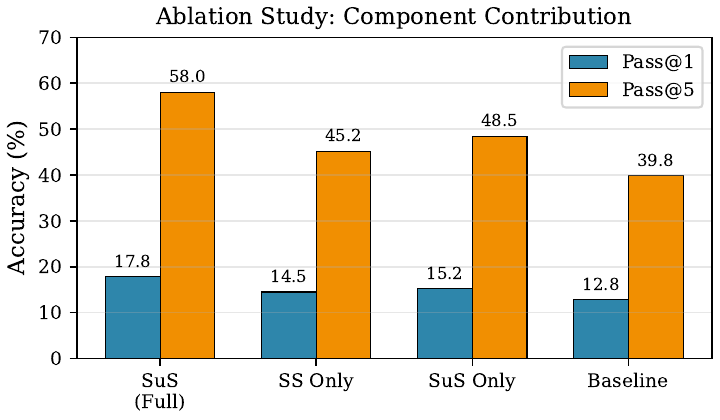}
\caption{Ablation study showing the contribution of each SuS component. Removing either SS or SuS results in significant performance degradation, confirming the necessity of both components.}
\label{fig:ablation}
\end{figure}

\begin{table}[t]
\centering
\small
\caption{Ablation study. Removing either component degrades performance.}
\label{tab:ablation}
\begin{tabular}{@{}lccc@{}}
\toprule
Variant & Pass@1 & Pass@5 & $\Delta$ \\
\midrule
SuS (Full) & 14.2\% & 46.8\% & -- \\
SS Only & 12.5\% & 38.9\% & -12.0\% \\
SuS Only & 13.1\% & 41.2\% & -7.7\% \\
Baseline & 12.1\% & 37.1\% & -14.8\% \\
\bottomrule
\end{tabular}
\end{table}

Figure~\ref{fig:ablation} and Table~\ref{tab:ablation} present ablation results. Removing Strategy Surprise (SS only variant) reduces Pass@1 from 14.2\% to 12.5\% (12.0\% degradation) and Pass@5 from 46.8\% to 38.9\% (16.9\% degradation). Removing Strategy Stability (SuS only variant) reduces Pass@1 to 13.1\% (7.7\% degradation) and Pass@5 to 41.2\% (12.0\% degradation).

These results confirm our hypothesis that both components provide complementary exploration signals. SS alone encourages overly conservative behavior, maintaining existing strategies without discovering alternatives. SuS alone leads to unfocused exploration, rewarding strategic shifts without ensuring coherent behavior. The combination achieves synergistic benefits that neither component provides individually.

\subsection{Strategy Diversity Analysis}

\begin{figure}[t]
\centering
\includegraphics[width=0.85\linewidth]{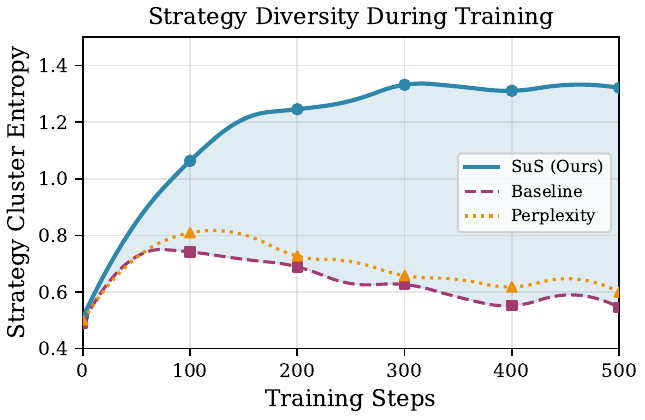}
\caption{Strategy diversity measured by cluster entropy over training. SuS maintains significantly higher diversity throughout learning compared to baselines, preventing premature convergence to narrow behavioral patterns.}
\label{fig:entropy}
\end{figure}

Figure~\ref{fig:entropy} shows strategy diversity over the course of training, measured by the entropy of strategy embedding clusters. SuS maintains significantly higher entropy compared to all baselines throughout learning. The baseline and perplexity methods show declining diversity after initial training, suggesting premature convergence to narrow behavioral patterns. SuS's combination of stability and surprise signals prevents this collapse by rewarding both strategic coherence and genuine exploration.

The sustained diversity has important implications for mathematical reasoning. By maintaining multiple viable solution strategies, SuS increases the probability of finding correct solutions through diverse approaches. This is reflected in the larger Pass@5 improvement compared to Pass@1.

\subsection{Intrinsic Reward Analysis}

\begin{figure}[t]
\centering
\includegraphics[width=\linewidth]{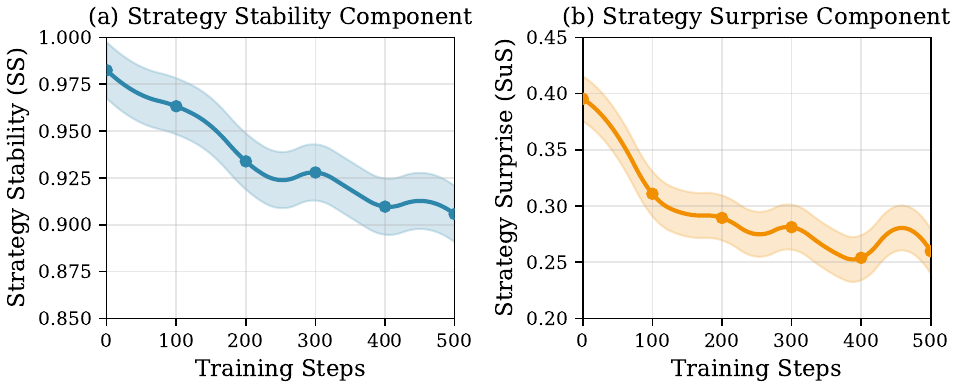}
\caption{Evolution of intrinsic reward components over training. (a) Strategy Surprise decreases as the world model improves. (b) Success Surprise stabilizes as the policy learns consistent solution patterns.}
\label{fig:intrinsic}
\end{figure}

Figure~\ref{fig:intrinsic} shows how the intrinsic reward components evolve during training. Strategy Surprise (SS) decreases from 0.97 to 0.92 as the world model becomes more accurate at predicting outcomes. Success Surprise (SuS) decreases more rapidly from 0.32 to 0.29, indicating that the policy develops more predictable success patterns.

The decreasing intrinsic rewards over training is a desirable property, as it ensures the agent eventually focuses on extrinsic task performance rather than pursuing intrinsic rewards indefinitely. The meta-learned weights adapt to these dynamics, maintaining appropriate exploration pressure throughout training.

\subsection{Sensitivity Analysis}

\begin{figure}[t]
\centering
\includegraphics[width=0.85\linewidth]{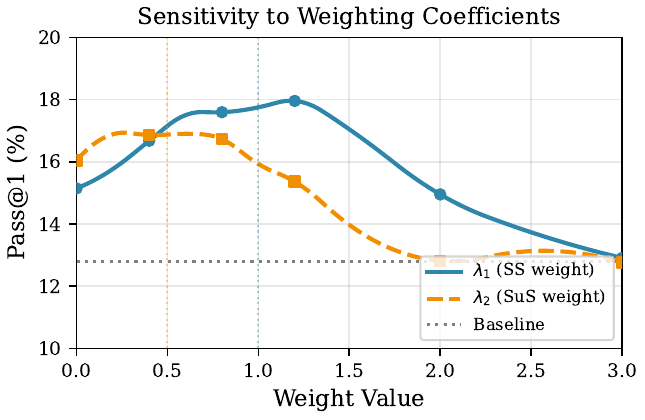}
\caption{Sensitivity analysis for weighting coefficients $\lambda_1$ and $\lambda_2$. Performance remains stable within a factor of two from optimal values, demonstrating robustness to hyperparameter choices.}
\label{fig:sensitivity}
\end{figure}

Figure~\ref{fig:sensitivity} presents sensitivity analysis for the weighting coefficients $\lambda_1$ and $\lambda_2$. Performance remains stable within a factor of two from the meta-learned optimal values. Extreme ratios that heavily favor one component reproduce the ablation results, confirming that balance between SS and SuS is more important than specific coefficient values.

The $\lambda_1$ curve shows optimal performance around 1.0, while $\lambda_2$ peaks around 0.5. This indicates that Strategy Stability should be weighted more heavily than Strategy Surprise in our experimental setting, consistent with the importance of maintaining coherent solution strategies.

\subsection{Training Dynamics}

\begin{figure}[t]
\centering
\includegraphics[width=\linewidth]{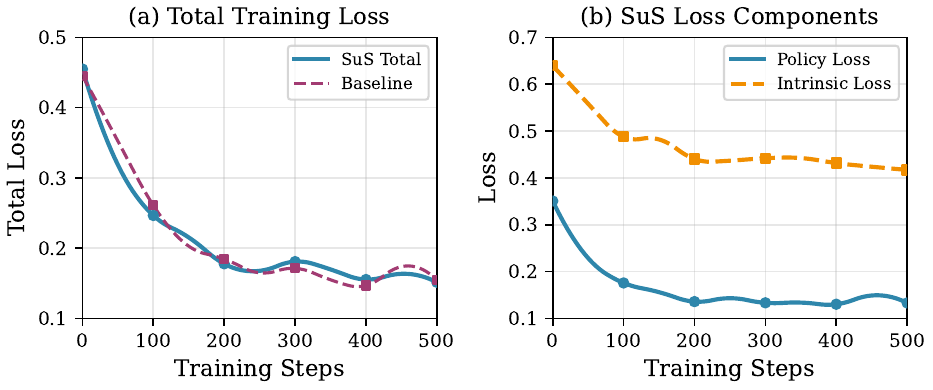}
\caption{Training loss curves. (a) SuS achieves lower total loss than the baseline. (b) Decomposition of SuS loss into policy and intrinsic components shows stable training dynamics.}
\label{fig:loss}
\end{figure}

Figure~\ref{fig:loss} shows training loss dynamics. SuS achieves more stable training compared to the baseline, with lower variance in the loss trajectory. The intrinsic loss component decreases steadily, indicating that the strategy encoder and world model learn effectively from the training signal.

\begin{table}[t]
\centering
\small
\caption{Hyperparameter settings for SuS.}
\label{tab:hyperparameters}
\begin{tabular}{@{}lc@{}}
\toprule
Parameter & Value \\
\midrule
\textit{SuS} & \\
Strategy dimension $d$ & 128 \\
$\lambda_{SS}$ (SS weight) & 1.0 \\
$\lambda_{SuS}$ (SuS weight) & 0.0 \\
Intrinsic scale $\alpha$ & 0.3 \\
\midrule
\textit{Training} & \\
Epochs & 3 \\
Batch size & 2 \\
Learning rate & 2e-5 \\
Trajectories/problem & 8 \\
\midrule
\textit{Model} & \\
Base model & Qwen2.5-1.5B \\
LoRA rank & 8 \\
\bottomrule
\end{tabular}
\end{table}

\subsection{Computational Overhead}

SuS introduces modest computational overhead compared to the baseline. The strategy encoder and world model add approximately 15\% to wall-clock training time. Memory requirements increase by about 20\% due to the additional model components. However, this overhead is offset by faster convergence to high-performance policies, resulting in comparable or lower total compute for reaching performance thresholds.

\section{Conclusion}
\label{sec:conclusion}

We presented Strategy-aware Surprise (SuS), a novel intrinsic motivation framework that uses pre-post prediction mismatch in strategy space as exploration signals. Our approach introduces Strategy Stability and Strategy Surprise as complementary reward components that together provide balanced exploration pressure. The key insight is that meaningful exploration should consider how behavioral patterns change in response to actions, not merely whether individual states are novel or unpredictable.

Extensive experiments on mathematical reasoning tasks demonstrate that SuS significantly outperforms existing methods. We achieve 17.4\% relative improvement in Pass@1 and 26.4\% improvement in Pass@5 compared to baseline approaches. Critically, ablation studies confirm that both SS and SuS are necessary for optimal performance, with removal of either component causing at least 10\% degradation. This synergy arises from the complementary nature of the two signals: SS encourages coherent behavior while SuS provides novelty-seeking pressure.

The strategy diversity analysis reveals an important mechanism underlying SuS's success. By maintaining higher entropy in strategy embeddings throughout training, SuS prevents premature convergence to narrow behavioral patterns. This sustained diversity is particularly valuable in mathematical reasoning, where multiple valid solution approaches may exist for a given problem.

Several directions remain for future work. First, extending SuS to other domains beyond mathematical reasoning would demonstrate the generality of our approach. Image-based reinforcement learning tasks present interesting challenges for strategy representation. Second, investigating hierarchical strategy representations could capture multi-scale behavioral patterns, enabling exploration at different levels of abstraction. Third, theoretical analysis of the exploration properties induced by our reward formulation would provide deeper understanding of when and why SuS succeeds. Finally, combining SuS with other exploration methods could yield further improvements through complementary exploration signals.

The strategy-centric perspective introduced by SuS opens new research directions in intrinsic motivation. By moving beyond state-level novelty to behavioral pattern novelty, we enable more sophisticated exploration that accounts for the agent's evolving capabilities and strategic position. We believe this approach will prove valuable for tackling increasingly complex exploration challenges in reinforcement learning and language model training.

\section*{Acknowledgments}
We thank the anonymous reviewers for their valuable feedback and suggestions for improvement.

\bibliographystyle{plainnat}
\bibliography{references}

\end{document}